# Orders-of-coupling representation with a single neural network with optimal neuron activation functions and without nonlinear parameter optimization

Sergei Manzhos[1], Manabu Ihara[2]

School of Materials and Chemical Technology, Tokyo Institute of Technology, Ookayama 2-12-1, Meguro-ku, Tokyo 152-8552 Japan

## Abstract

Representations of multivariate functions with low-dimensional functions that depend on subsets of original coordinates (corresponding of different orders of coupling) are useful in quantum dynamics and other applications, especially where integration is needed. Such representations can be conveniently built with machine learning methods, and previously, methods building the lower-dimensional terms of such representations with neural networks [e.g. Comput. Phys. Comm. 180 (2009) 2002] and Gaussian process regressions [e.g. Mach. Learn. Sci. Technol. 3 (2022) 01LT02] were proposed. Here, we show that neural network models of orders-of-coupling representations can be easily built by using a recently proposed neural network with optimal neuron activation functions computed with a first-order additive Gaussian process regression [arXiv:2301.05567] and avoiding non-linear parameter optimization. Examples are given of representations of molecular potential energy surfaces.

[1] E-mail: manzhos.s.aa@m.titech.ac.jp
[2] E-mail: mihara@chemeng.titech.ac.jp

# 1 Introduction

Construction of high-dimensional functions $f(x), x \in R^D$ from samples and their use in applications can be difficult as $D$ becomes high, especially when integration is required. In theoretical chemistry, for example, variational methods to solve the Schrödinger equation critically depend on the calculation of integrals [1,2]. The size of a numeric integration grid grows exponentially in the case of a direct product grid. Approaches such as Monte Carlo integration [3] or use of sparse grids such as Smolyak grids [4] alleviate but do not dispense with the problem. In-substance solutions include sums-of-products (SOP) and orders-of-coupling representations, which allow dealing only with low-dimensional integrals. SOP representations have the form

$$f(x) = \sum_{n=1}^{N} \prod_{i=1}^{D} f_{kn}(x_i)$$

(1.1)

Algorithms have been proposed to cast $f(x)$ in a SOP form [5,6], including machine-learning algorithms based on neural networks (NN) [7] using suitable (non-sigmoid) neuron activation functions [8–11]. Orders of coupling representations have the form

$$f(x) \approx f_0 + \sum_{i=1}^{D} f_i(x_i) + \sum_{1 \leq i < j \leq D} f_{ij}(x_i, x_j) + \cdots + \sum_{\{i_1 i_2 \ldots i_d\} \in \{12 \ldots D\}} f_{i_1 i_2 \ldots i_d}(x_{i_1}, x_{i_2}, \ldots, x_{i_d})$$

(1.2)

This representation was formalized under the name of HDMR (high-dimensional model representation) in a series of papers by Rabitz et al. [12–15]. Multi-body and N-mode [16] representations well-known in chemistry and physics are particular cases of HDMR. When $d = D$, this representation is exact, when $d < D$, it is approximate. This approximation is justified by the fact that in many applications, a rather low maximum included order of coupling $d$ ($d$ = 2…4) is sufficient to achieve the desired accuracy [12,17–21]. Also, with finite data, higher-order coupling terms may not be recoverable [17,19]. Note that in the general case (what Rabitz et al. called random-sampling (RS) HMDR, where "random sampling" should be understood as the possibility to work with any desired distribution of samples of $f(x)$ rather than randomness), the component functions $f_{i_1 i_2 \ldots i_d}(x_{i_1}, x_{i_2}, \ldots, x_{i_d})$ are not slices of $f(x)$ along $d$-dimensional hyper-surfaces (as in the N-mode approximation) [12,13,15], and all component functions are determined from one and the same set of samples of $f(x)$. In the original formulation of RS-HDMR, different component functions are mutually orthogonal, and $d$-dimensional component functions are computed with $(D-d)$-dimensional integrals, which may be costly [12,15,22].

Some of us proposed a modified version of HDMR with machine-learned (ML) component functions [17–20,23–27]. This dispenses with the need to wield $(D-d)$-dimensional integrals and otherwise makes for an easy-to-use approach if one does not enforce orthogonality of component functions. Lower-order terms can also be lumped into the highest-order $d$-dimensional terms, resulting in an approximation

$$f(x) \approx \sum_{\{i_1 i_2 \ldots i_d\} \in \{12 \ldots D\}} f^{ML}_{i_1 i_2 \ldots i_d}(x_{i_1}, x_{i_2}, \ldots, x_{i_d})$$

(1.3)

Versions with neural network based component functions (HDMR-NN) [17,20,23–26] and Gaussian process regression (GPR) based component functions (HDMR-GPR) [18,19,27] have been proposed.

HDMR-NN can be envisioned as separate NN instances for each summand of Eq. (1.3) fitted self-consistently to the difference of the samples of $f(x)$ and the sum of all other component functions, as is done in an available implementation [20] (with a similar idea also used in an available HDMR-GPR implementation [18]),

$$f_{k_1 k_2 \ldots k_d}(x_{k_1}, x_{k_2}, \ldots, x_{k_d}) = f(x) - \sum_{\substack{\{i_1 i_2 \ldots i_d\} \in \{12 \ldots D\} \\ \{i_1 i_2 \ldots i_d\} \neq \{k_1 k_2 \ldots k_3\}}} f_{i_1 i_2 \ldots i_d}(x_{i_1}, x_{i_2}, \ldots, x_{i_d})$$

(1.4)

Alternatively, one can imagine a single NN with an architecture (connectivity) realizing HDMR [17]. This is a leaner, compacter, more esthetically pleasing but a much CPU-costlier approach because of the cost of non-linear NN parameter optimization which becomes much higher if all terms are included into one NN. To the best of our knowledge, there is as of now no implementation of a single NN realizing HDMR. Here, we offer such implementation by dispensing with non-linear optimization altogether.

A traditional single hidden layer feed-forward neural network approximation of $f(x)$ has the form

$$f(x) = c_{out}\sigma_{out}\left(\sum_{i=1}^{N} c_i \sigma_i(\mathbf{w}_i x + b_i) + b_{out}\right)$$

(1.5)

where $\sigma_i$ are one-dimensional neuron activation functions, $\mathbf{w}_i$ are the weights, and $b_i$ are the biases of the so-called hidden layer. The output neuron $\sigma_{out}$ (which must be monotonic), the output bias $b_{out}$, and the output wight $c_{out}$ are omitted in the following, as they can be subsumed into the left-hand side without loss of generality. Usually, one and the same $\sigma(x)$ is used for all neurons, i.e. $\sigma_i(x) = \sigma(x)$. One often uses sigmoid activation functions for which earlier versions of the universal approximator theorem were developed [28,29]. Weights and biases are non-linear parameters whose optimization is a major component of the CPU cost of training a NN model. Non-linear optimization also exposes NNs to the problems of local minima and overfitting which may be worse than with a linear model when data are few and parameters are many. The single-hidden layer NN is already a universal approximator [30,31]. This means that for any predefined $\delta > 0$, there exists a finite $N$ such that $|f(x) - \sum_{n=0}^{N} c_n \sigma(\mathbf{w}_n x + b_n)| < \delta$. The universal approximator theorem does not concern itself with the issues stemming from a finite data set size (finite sampling density) and associated overfitting. In this respect, the use of HDMR has the advantage that lower-dimensional terms can be reliably constructed with fewer data [12,18,19,32].

In Ref. [33], we proposed an approach to build a single hidden layer NN that dispenses with non-linear parameter optimization. We recognized that Eq. (1.5) has the same form as an additive model

$$f(\boldsymbol{x}) = \sum_{i=1}^{N} c_i \sigma_i(y_i) = \sum_{i=1}^{N} f_i(y_i)$$

(1.6)

in redundant coordinates $\boldsymbol{y}$ that linearly depend on $\boldsymbol{x}$, $\boldsymbol{y} = \boldsymbol{Wx} + \boldsymbol{b}$, with $\boldsymbol{W}$ a matrix of all $\boldsymbol{w}_i$ and $\boldsymbol{b}$ a vector of all $b_i$. We showed that one can use a rule-based definition of $\boldsymbol{W}$ to form $\boldsymbol{y}$ and to use a first-order additive Gaussian process (1$d$-HDMR-GPR) [27,34] to build optimal neuron activation functions $\sigma_i(y_i)$ (component functions $f_i(x_i)$) for each $y_i$ [33]. Note that neuron activation functions can be different for different neurons and they can be any smooth non-linear functions [35]; the component functions of an additive GPR satisfy this condition as long as the GPR kernel is smooth. The biases are not needed as they are automatically taken care of by the form of $\sigma_i(y_i)$. In the present article, we show that an NN with an architecture realizing HDMR is easily obtained by modifying the rules that define $\boldsymbol{W}$. As an application, we construct orders of coupling representations of potential energy surfaces (PES) of water and formaldehyde molecules in the spectroscopically relevant region. This application and these PESs are chosen as test cases because the required and obtainable accuracy is well understood from prior literature [36,37], and because orders-of-coupling representations of these PESs were previously obtained with other approaches [19,27], to which the results of the present work can be compared.

## 2 Methods

Starting from the original coordinates $\boldsymbol{x}$ and a chosen order of coupling $d$, redundant coordinates $\boldsymbol{y} = \boldsymbol{Wx}$ are generated as $y_i = \boldsymbol{x}^T \boldsymbol{s}_i$ where $\boldsymbol{s}_i$ is a vector whose $d$ elements are taken from a $d$-dimensional pseudorandom Sobol sequence [38], and the other $(D - d)$ elements are zero. The order of the $d$ elements determines the selection of a particular coupling term dependent on $(x_{i_1}, x_{i_2}, \ldots, x_{i_d})$. The elements of the Sobol sequence determine $\boldsymbol{W}$. We include the original coordinate vector $\boldsymbol{x}$ into $\boldsymbol{y}$ (i.e. the newly formed $y_i$ are added to the list of features); this corresponds to having within $(x_{i_1}, x_{i_2}, \ldots, x_{i_d})$ cases where some $x_{i_k}$ are zero. One can select a desired number $N_{i_1 i_2 \ldots i_d}$ of $y_n = y_n(x_{i_1}, x_{i_2}, \ldots, x_{i_d})$ for each coupling term. The resulting approximation is then an orders-of-coupling representation

$$f(x) = \sum_{\{i_1 i_2 \ldots i_d\} \in \{12 \ldots D\}} f_{i_1 i_2 \ldots i_d}(x_{i_1}, x_{i_2}, \ldots, x_{i_d}) = \sum_{i_1 i_2 \ldots i_d} \sum_{j=1}^{N_{i_1 i_2 \ldots i_d}} \sigma_{j, i_1 i_2 \ldots i_d}(x_{i_1}, x_{i_2}, \ldots, x_{i_d})$$

(2.1)

Eq. (2.1) has the form of Eq. (1.6) where the coefficients $c_n$ are subsumed into $\sigma_i$, $y_i = w_i x$, and $N_{i_1 i_2 \ldots i_d}$ rows $w_i$ of matrix $W$ contain $d$ non-zero elements, for all selections of $d$ components $(x_{i_1}, x_{i_2}, \ldots, x_{i_d})$ of $D$. That is, this is a HMDR-NN where $f_{i_1 i_2 \ldots i_d}(x_{i_1}, x_{i_2}, \ldots, x_{i_d})$ are single hidden layer NNs with customized neurons. In the following section, for simplicity, we use the same $N_{i_1 i_2 \ldots i_d}$ for all coupling terms, which is sufficient for our purpose. If desired, some permutations (coupling terms) can be omitted, e.g. if they are known or found to be unimportant [18].

The shapes of the functions $\sigma_i$ are found by applying a first-order additive GPR (HDMR-GPR) [27] to fit $f(y)$. The additive GPR is implemented in Matlab with a custom kernel of additive form,

$$k(x, x') = \sum_{i=1}^{D} k_i(x_i, x'_i)$$

(2.2)

where the components $k_i(x_i, x'_i)$ are one-dimensional square exponential kernels

$$k_i(x_i, x'_i) = exp\left(-\frac{(x_i - x'_i)^2}{2l^2}\right)$$

(2.3)

As the features $y_i$ were scaled to unit cube before regression, we used a single value of $l$ for all $i$. The value of $l$ is about 0.45 for $H_2O$ and 0.9 for $H_2CO$, respectively. No significant changes are observed in the results for small variations of $l$ around these values. The noise parameter was set to $1 \times 10^{-6}$.

In the following, we will call the method for brevity HDMR-NN(GPR). We apply the method to form orders-of-coupling representations of the PESs of $H_2O$ and $H_2CO$. For the description of the datasets and information about the applications of these PESs, specifically in computational spectroscopy [1], see Ref. [37] for $H_2O$ and Ref. [39] for $H_2CO$. The three-dimensional dataset for $H_2O$ PES contains 10,000 data points sampled in Radau coordinates [40]. The values are sampled from the analytic PES of Ref. [41] and range 0-20,000 $cm^{-1}$. The six-dimensional dataset for $H_2CO$ is sampled from the analytic PES of Ref. [42] and contains 120,000 data points in bond coordinates (intra-molecular bonds and angles) with values of potential energy ranging 0-17,000 $cm^{-1}$. Both datasets are available in the supporting information of Ref. [18].

The data points sample the PES around the equilibrium geometry over a sufficiently wide range of coordinates to include anharmonic regions and allow calculations of hundreds of vibrational levels [37]. We use a randomly selected subset of $M = 1000$ points for both $H_2O$ and $H_2CO$ for training and 10,000 points (for $H_2CO$) or the remaining 10,000-$M$ points (for $H_2O$) for testing. The test set is thus much larger than the training set to reliably gauge a global quality of the regression. We performed fits with other values of M, they lead to the same conclusions. We therefore present here results with $M = 1000$. Due to the random nature of the choice of the training points from the full dataset, there is a slight variation of the error from run to run which does not significantly affect the results. We take the best value from 3 runs.

## 3  Results

In Figure 1, we show test and train rmse values achieved with HDMR-NN(GPR) of different orders of coupling $d$ and using different numbers of neurons $N$ when fitting the PES of $H_2O$. The corresponding correlation plots between fitted and exact values are shown in Figure 2, Figure 3, and Figure 4. In those figures, we also show the relative importance of component functions $f_i$ and their shapes for the most important (by magnitude) functions. In the case of this three-dimensional $H_2O$ dataset, only the $d = 2$ case is non-trivial. The $d = 1$ case is the same as the first order additive GPR, no redundant $y_i$ are added. The $d = 3$ case is a full-dimensional NN with first-order HDMR-GPR neuron activation functions, as per Ref. [33].

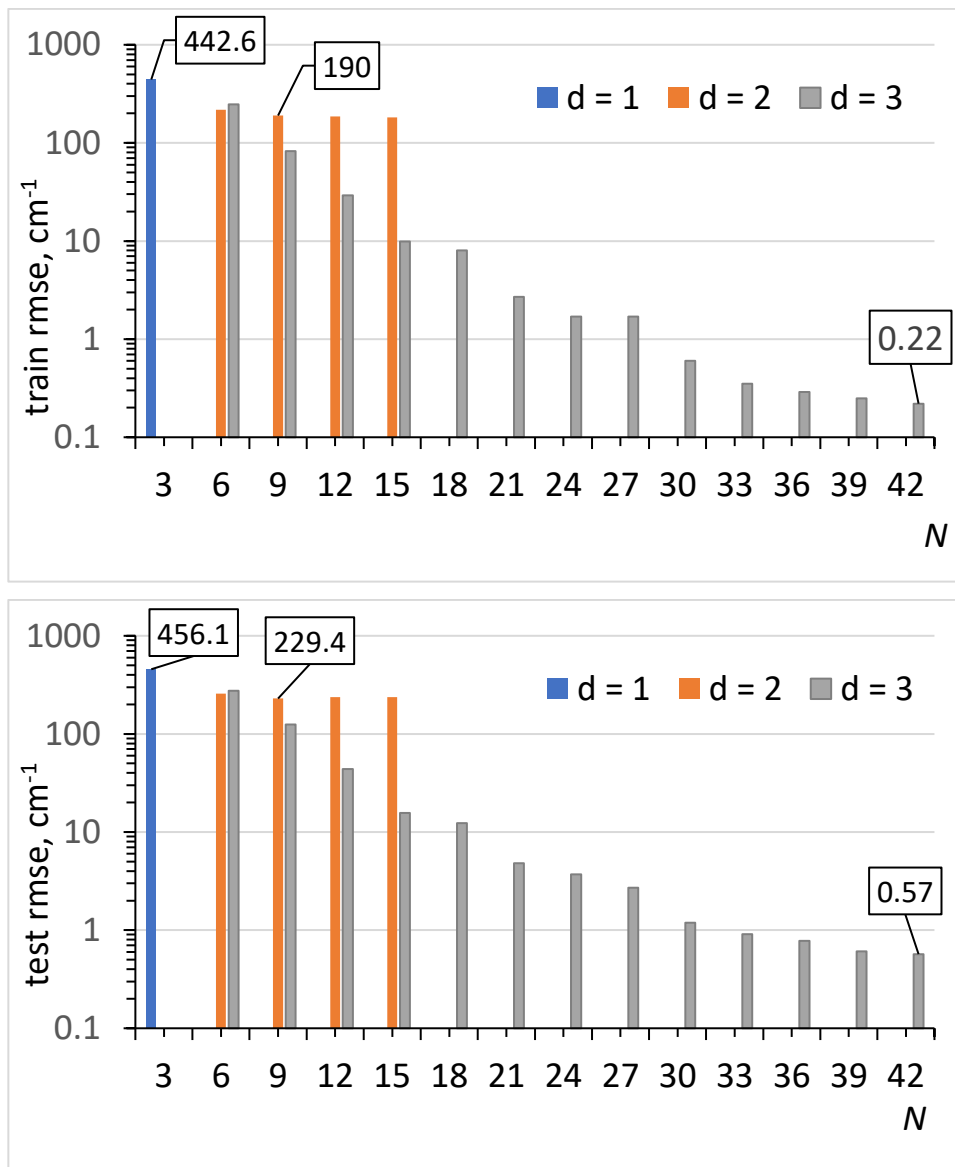

Figure 1. Train (top) and test (bottom) rmse values when fitting the PES of $H_2O$ with HDMR-NN(GPR) of different orders of coupling $d$, when using different numbers of neurons $N$. Note the logarithmic scale.

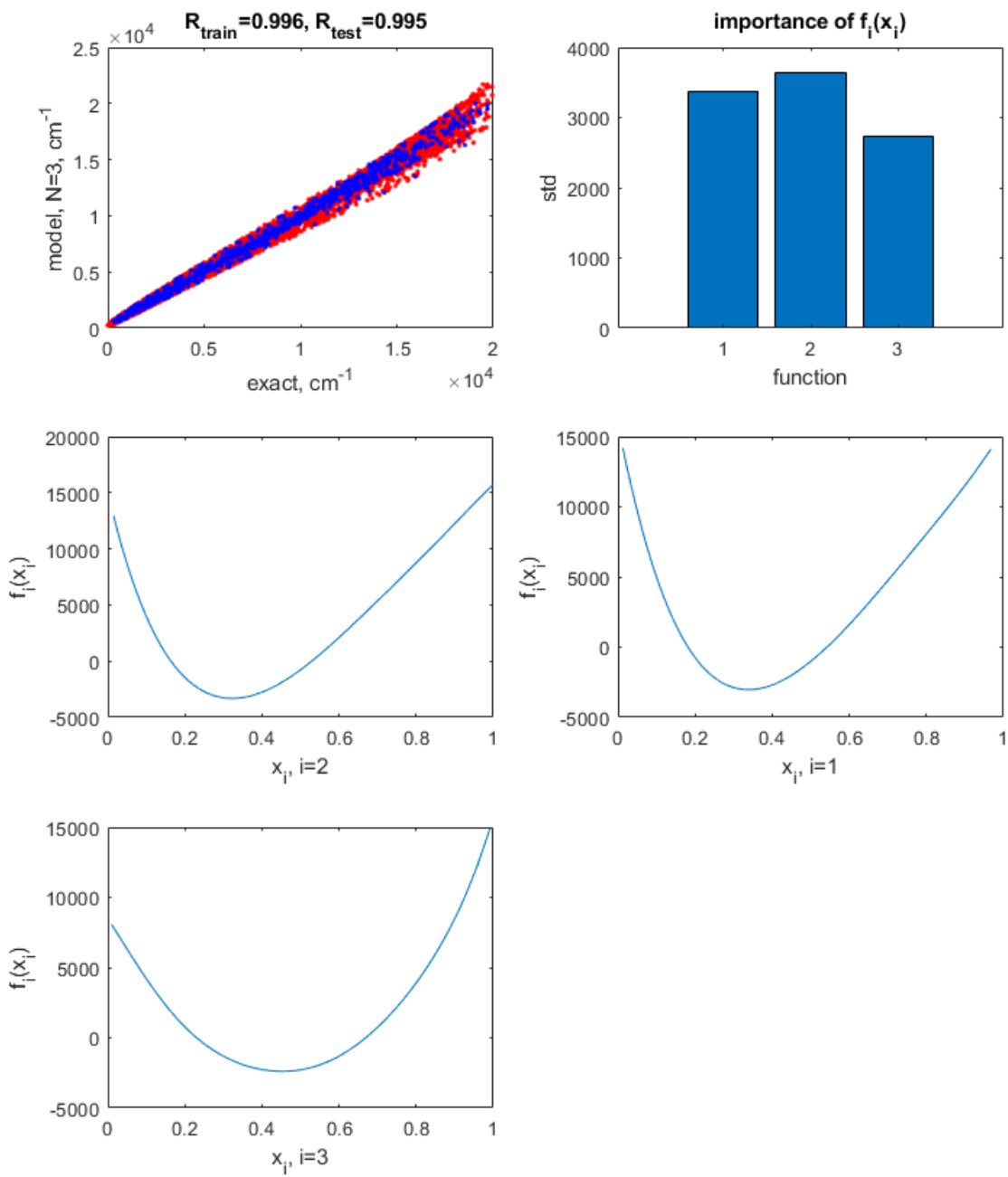

Figure 2. The results of fitting a first-order ($d = 1$) additive model to the PES of $H_2O$ with $N = 3$ terms. Top left: the correlation plot between the exact and fitted values (blue: training set, red: test set). The correlation coefficients for the train and test sets are given on the plot. Top right: the relative importance of terms by standard deviation ("std"). The following three plots are the shapes of the component functions $f_i(x_i)$.

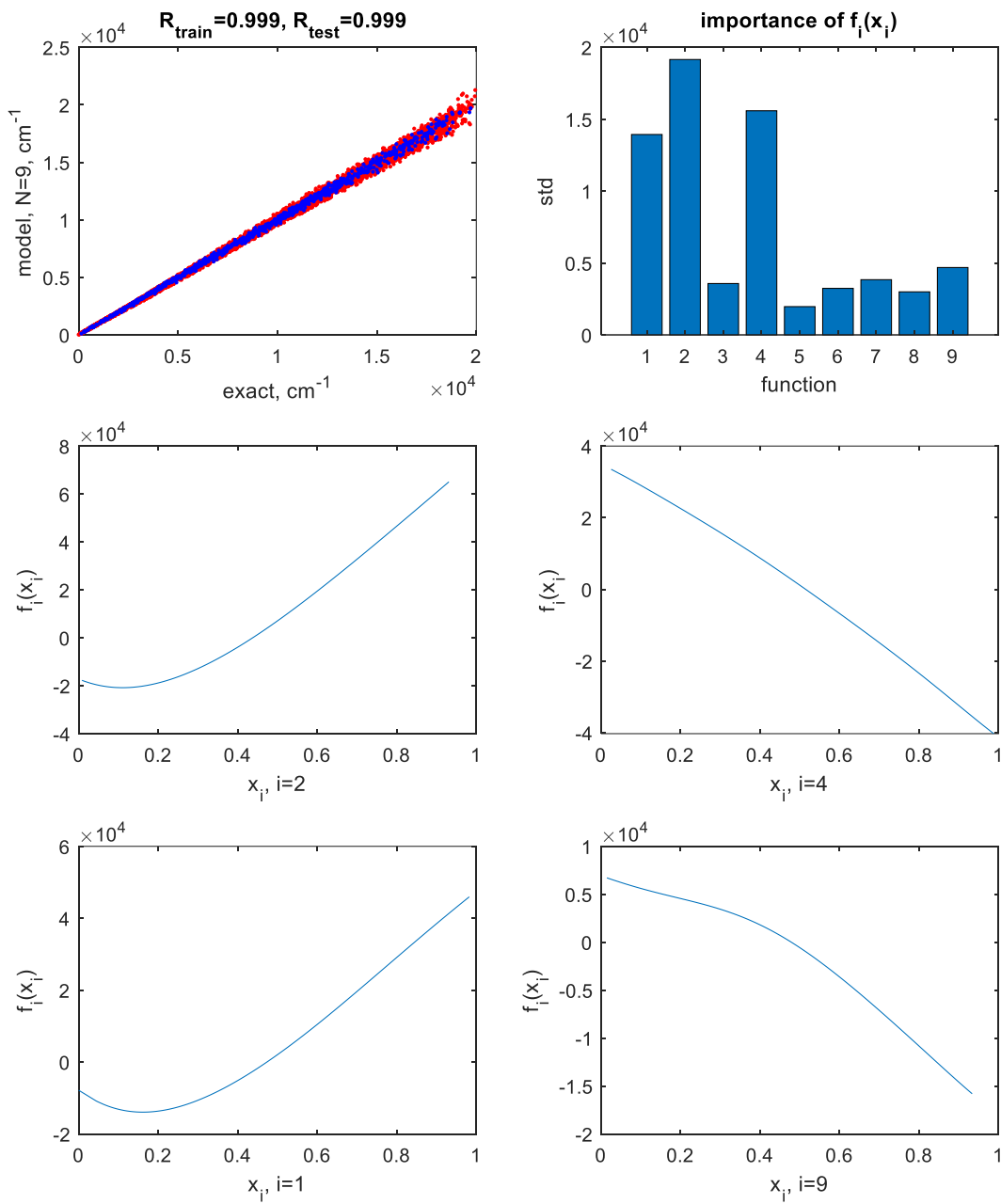

Figure 3. The results of fitting a second-order ($d = 2$) HDMR-NN(GPR) model to the PES of $H_2O$ with $N$ terms-neurons. Top left: the correlation plot between the exact and fitted values (blue: training set, red: test set). The correlation coefficients for the train and test sets are given on the plot. Top right: the relative importance of terms by standard deviation ("std"). The following four plots are the shapes of the four most important component functions $f_i(x_i)$.

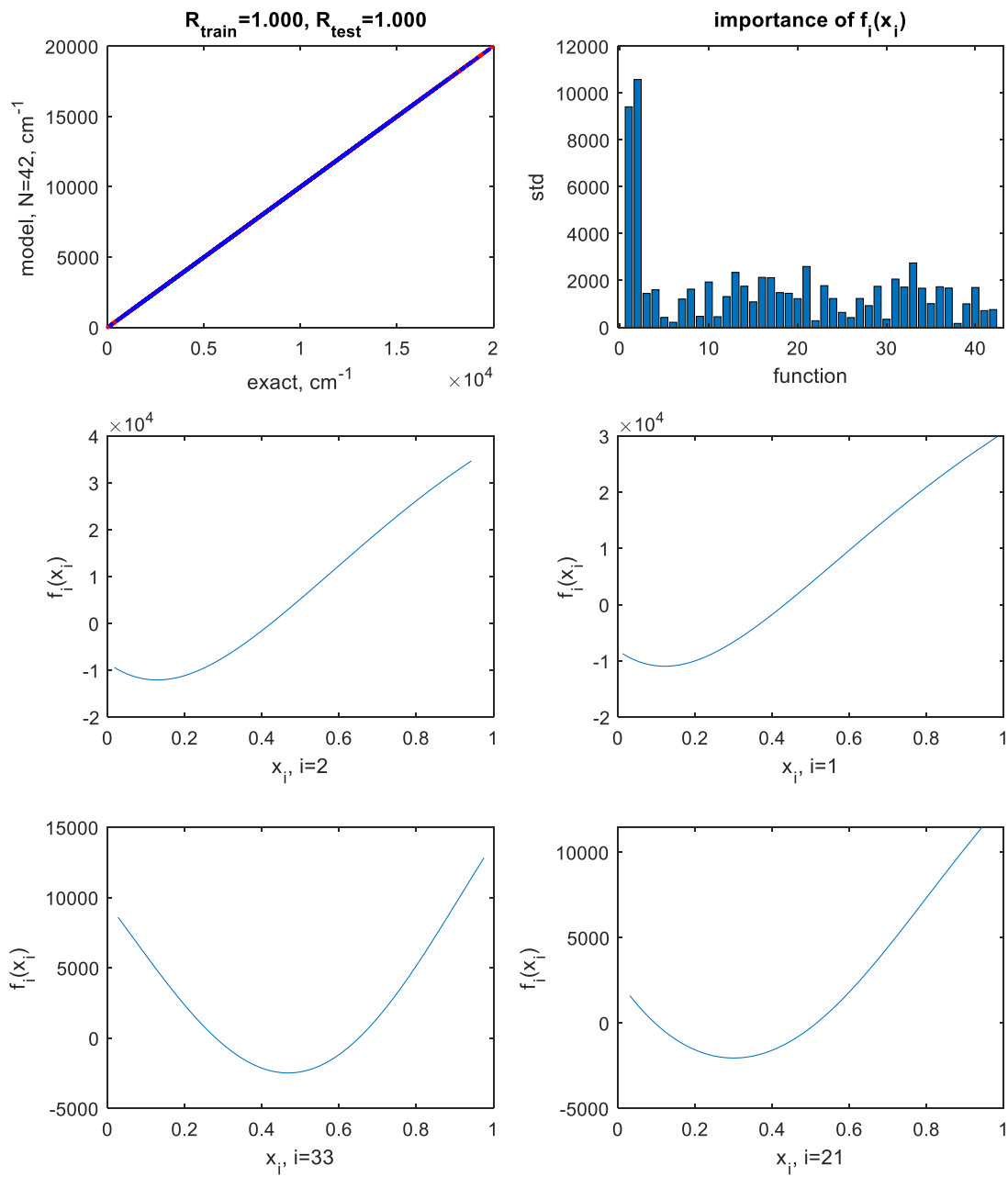

Figure 4. The results of fitting a third-order (full dimensional) HDMR-NN(GPR) model to the PES of $H_2O$ with *N* terms-neurons. Top left: the correlation plot between the exact and fitted values (blue: training set, red: test set). The correlation coefficients for the train and test sets are given on the plot. Top right: the relative importance of terms by standard deviation ("std"). The following four plots are the shapes of the four most important component functions $f_i(x_i)$.

The rmse values can be compared to those obtained for the same dataset and the same number of training points in Ref. [18] with HDMR-GPR: with $M = 1000$, they obtained a test error of 462.5 and 259.1 cm$^{-1}$ for $d = 1$ and $d = 2$, respectively. They also reported a test rmse with a plain GPR model (corresponding to $d = 3$) of 1.9 cm$^{-1}$ but with 500 training points. We obtain, for $d = 2$, 0.57 cm$^{-1}$ with $M = 1000$ and 1.33 cm$^{-1}$ with $M = 500$. The method therefore performs well in spite of the absence of any non-linear parameter optimization. The minimum achievable test rmse for $d = 1$ and $d = 2$ is limited by the neglect of coupling and rapidly saturates with $N$ (for $d = 1$, $N$-dependence has no meaning) with similar values for train and test errors, whereby the train set error cannot be reduced. The minimum achievable test rmse for $d = 3$ is limited by the density of data and saturates at $N$ of about 40, with a wider divergence between train and test errors, with a possibility of further reduction of the train error by increasing $N$.

We now test HDMR-NN(GPR) on the six-dimensional formaldehyde data. Figure 5 and Figure 6 show train (Figure 5) and test (Figure 6) set rmse values obtained when fitting the PES of H$_2$CO with HDMR-NN(GPR) of different orders of coupling $d$ using different numbers of neurons $N$. The corresponding correlation plots between fitted and exact values, the relative importance of component functions $f_i$ and their shapes for the most important (by magnitude) functions are shown in the Supporting Information. Similar to the case of H$_2$O, the fit accuracy at low $d$ is limited by the neglect of coupling, while the accuracy of higher-order terms is limited by the amount of data. In this example, the dimensionality is sufficiently high and the density of sampling is sufficiently low to see that highest-order coupling terms cannot be recovered [17,19]: the lowest *test* set error is obtained at $d = 4$. The test set error of the fifth-order model is similar, while the best test set error of the full-dimensional model is noticeably higher. The test set error of the third-order model ($d = 3$) is only slightly higher than that of the fourth-order model. For $d \geq 3$, the correlation plots are practically diagonal, while the much lower accuracy of the first- and second-order models can be visually appreciated from them (see Supporting Information).

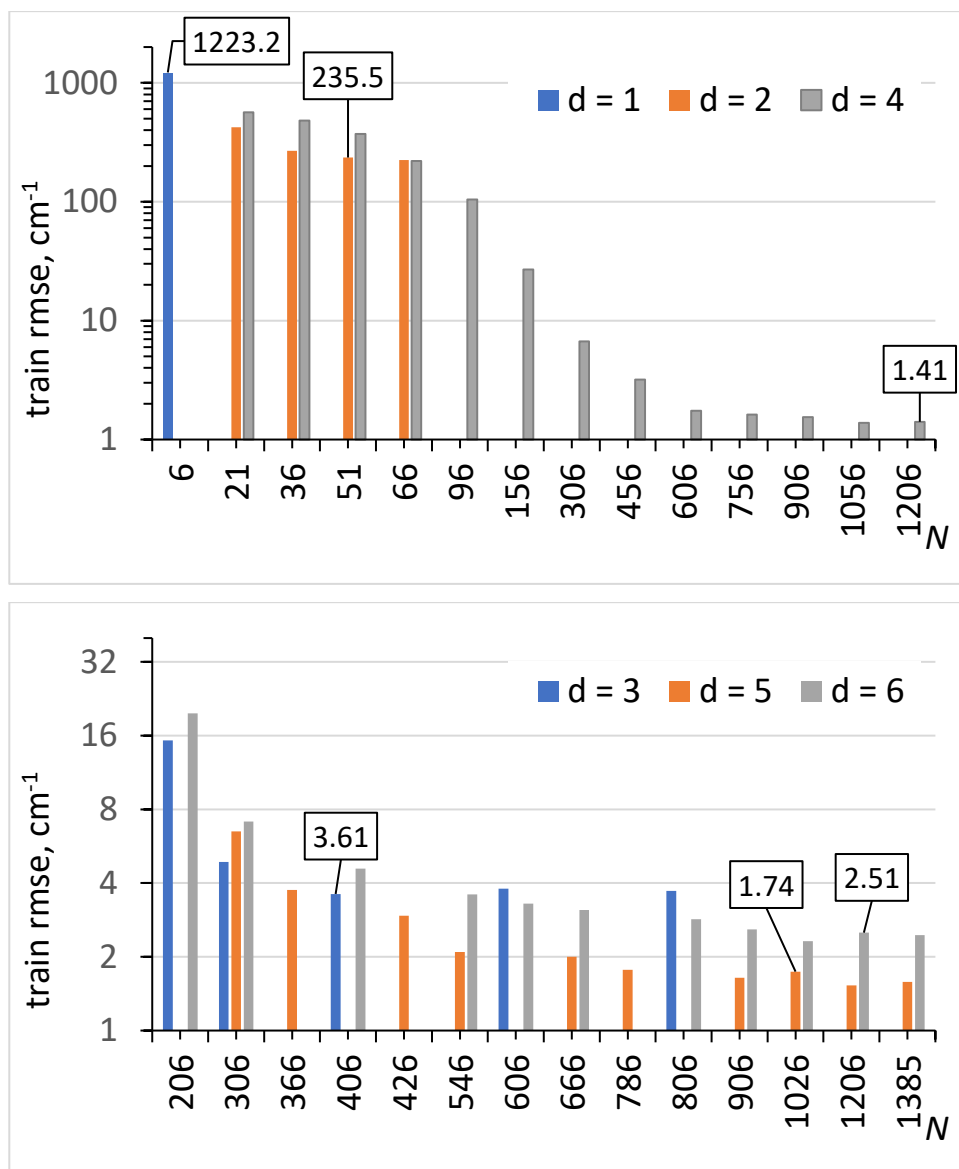

Figure 5. Train set rmse values when fitting the PES of $H_2CO$ with HDMR-NN(GPR) of different orders of coupling $d$, when using different numbers of neurons $N$. Note the logarithmic scale.

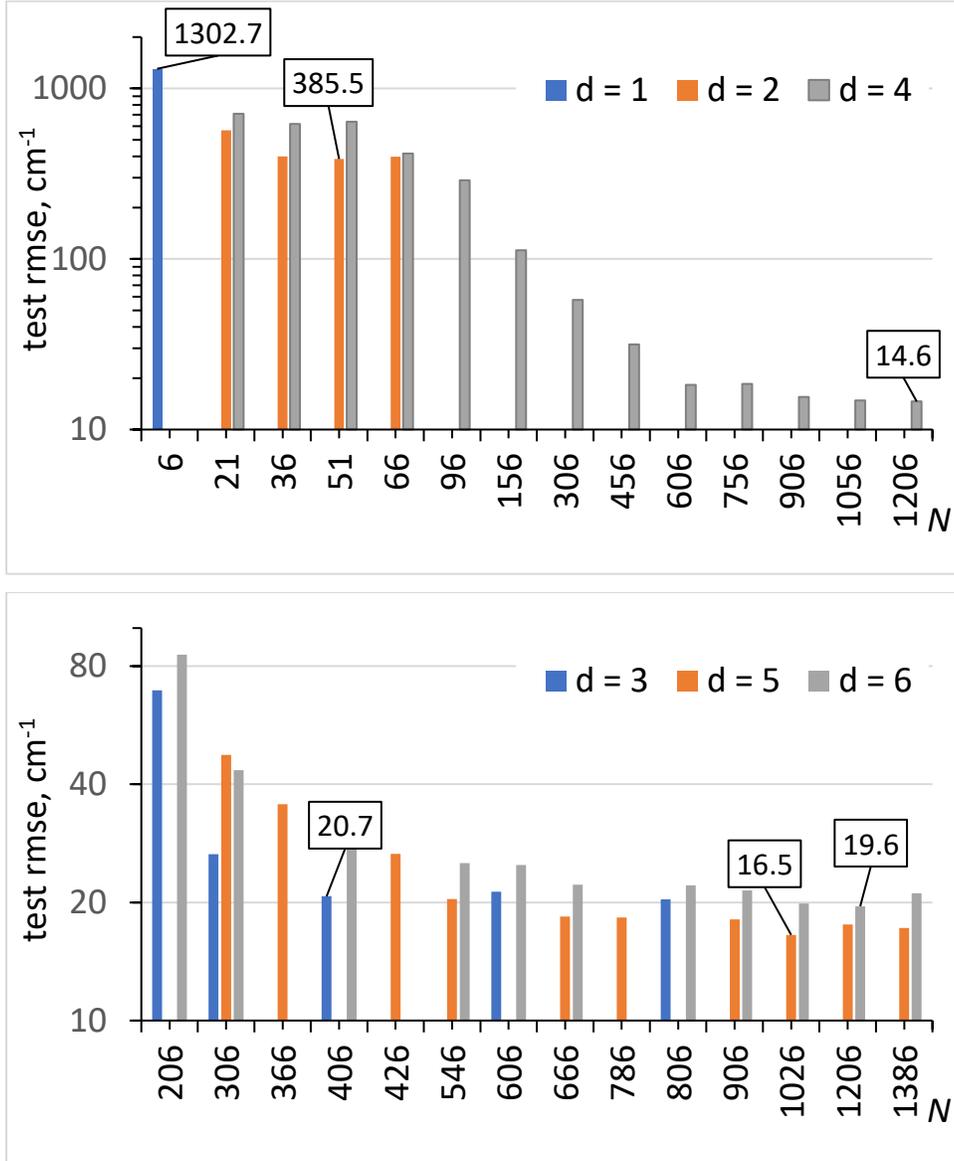

Figure 6. Test set rmse values when fitting the PES of $H_2CO$ with HDMR-NN(GPR) of different orders of coupling $d$, when using different numbers of neurons $N$. Note the logarithmic scale.

The lowest achievable test set rmse values for each $d$ are summarized in Table 1. The table also compares with the values at different $d$ achieved with HDMR-GPR in Ref. [19]. The HDMR-NN(GPR) method achieves a similar to slightly better performance. What is important here is not any improvement vs other (RS-)HDMR techniques (which all are expected to lead to roughly similar performance) but the simplicity of the concept with which it is achieved here, notably, the absence of any non-linear optimization in a single NN realizing HDMR via a rule-based definition of weights.

Table 1. Comparison of test set rmse values achieved with different orders $d$ of HDMR-NN(GPR) (this work) and HDMR-GPR (Ref. [19]), when fitting the PES of $H_2CO$ to the same data.

| Order of coupling $d$ | No. of coupling terms | Test rmse, cm$^{-1}$ (this work) | Test rmse, cm$^{-1}$ (Ref. [19]) |
|---|---|---|---|
| 1 | 6 | 1302.7 | 1315.6 |
| 2 | 15 | 385.5 | 410.0 |
| 3 | 20 | 20.4 | 29.1 |
| 4 | 15 | 14.6 | 16.1 |
| 5 | 6 | 16.5 | 14.4 |
| 6 | 1 | 19.6 | 23.8 |

## 4 Conclusions

We have shown that a representation of a multivariate function $f(x)$ with lower-dimensional functions – coupling terms dependent on subset of components of $x$ can be efficiently done with a rule-based definition of weights of a single-hidden later neural network with customized neuron activation functions for each neuron. No non-linear optimization is done, and the neuron activation functions are made optimal for the respective rule-based neuron arguments by using additive Gaussian process regression. The method is tested by building RS-HDMR type orders-of-coupling representations of potential energy surfaces of the water (3D) and formaldehyde (6D) molecules. The results showed similar-to-better performance vs previous HDMR constructions on the same data. What we believe is important is not any improvement vs other (RS-)HDMR techniques (which all are expected to lead to similar performance) but the simplicity of the concept with which an HDMR representation is achieved here, namely, in a single NN realizing HDMR via a rule-based definition of weights and without any non-linear optimization.

The present results tother with those of Ref. [33] indicate that there is value in developing neural networks with parameters determined by rules rather than the costly non-linear optimization, including NNs with structured connectivity realizing desired properties (such as HDMR). With rule-based NN weights, we did not observe any substantial increase in test set error as the number of neurons is increased (contrary to a typical behavior of a standard NN), indicating that static, rule-based parameters are useful not only to reduce the CPU cost but also to improve stability and suppress overfitting. In this work, we used a simple and general rule; further research into different rules for setting NN weights might be fruitful.

## 5  Acknowledgements

This work was supported by JST-Mirai Program Grant Number JPMJMI22H1, Japan.

## 6  Data availability statement

The Matlab code used in the presented calculations are available from the corresponding author upon reasonable request. The data used are available in the supporting information of Ref. [18].

# Supporting Information

# Orders-of-coupling representation with a single neural network with optimal neuron activation functions and without nonlinear parameter optimization

Sergei Manzhos[1], Manabu Ihara[2]

School of Materials and Chemical Technology, Tokyo Institute of Technology, Ookayama 2-12-1, Meguro-ku, Tokyo 152-8552 Japan

[1] E-mail: manzhos.s.aa@m.titech.ac.jp
[2] E-mail: mihara@chemeng.titech.ac.jp

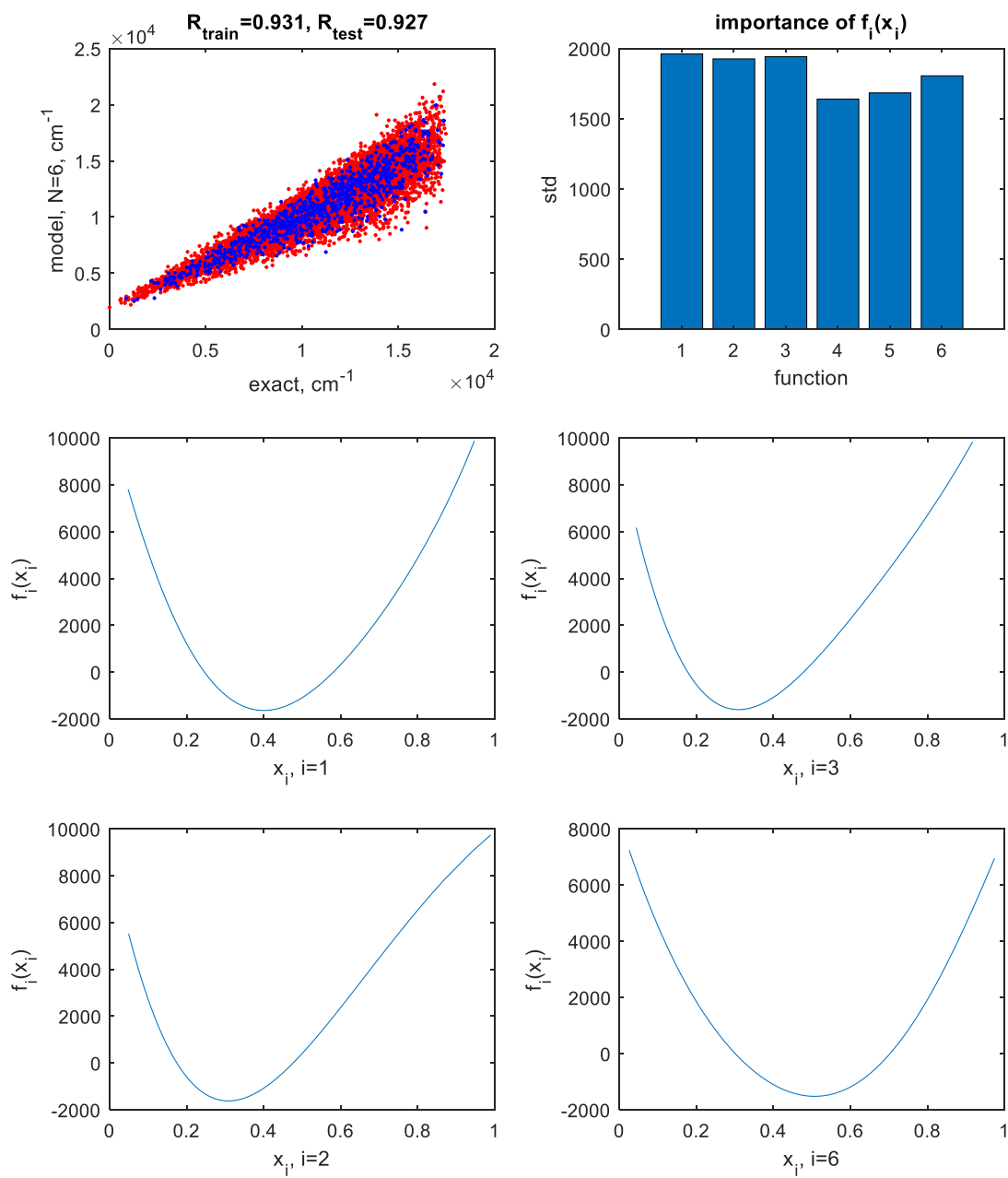

Figure 1. The results of fitting a first-order ($d = 1$) HDMR-NN(GPR) model to the PES of $H_2CO$ with $N$ terms-neurons. Top left: the correlation plot between the exact and fitted values (blue: training set, red: test set). The correlation coefficients for the train and test sets are given on the plot. Top right: the relative importance of terms by standard deviation ("std"). The following three plots are the shapes of the first four most important component functions $f_i(x_i)$.

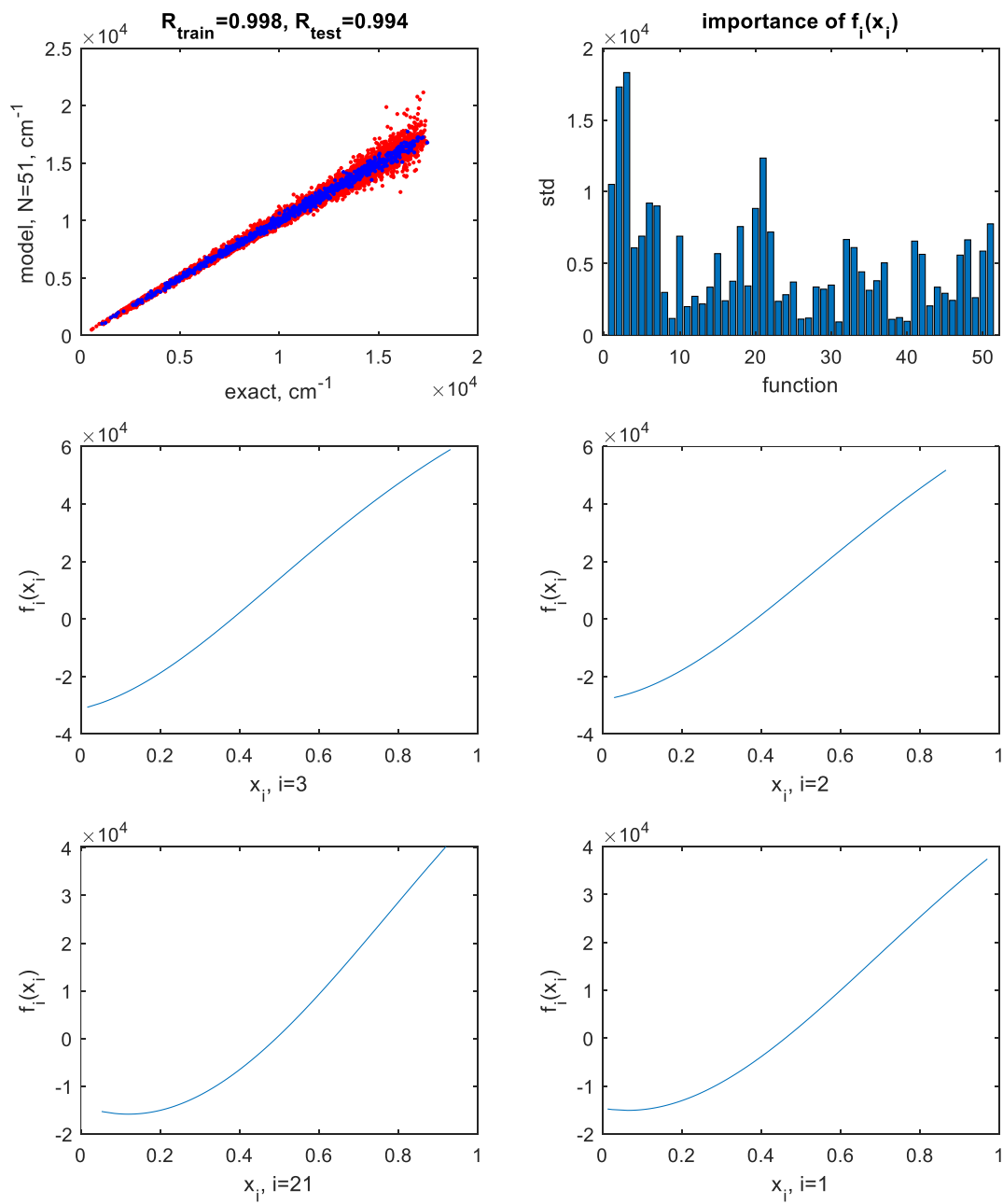

Figure 2. The results of fitting a second-order ($d = 2$) HDMR-NN(GPR) model to the PES of $H_2CO$ with *N* terms-neurons. Top left: the correlation plot between the exact and fitted values (blue: training set, red: test set). The correlation coefficients for the train and test sets are given on the plot. Top right: the relative importance of terms by standard deviation ("std"). The following three plots are the shapes of the first four most important component functions $f_i(x_i)$.

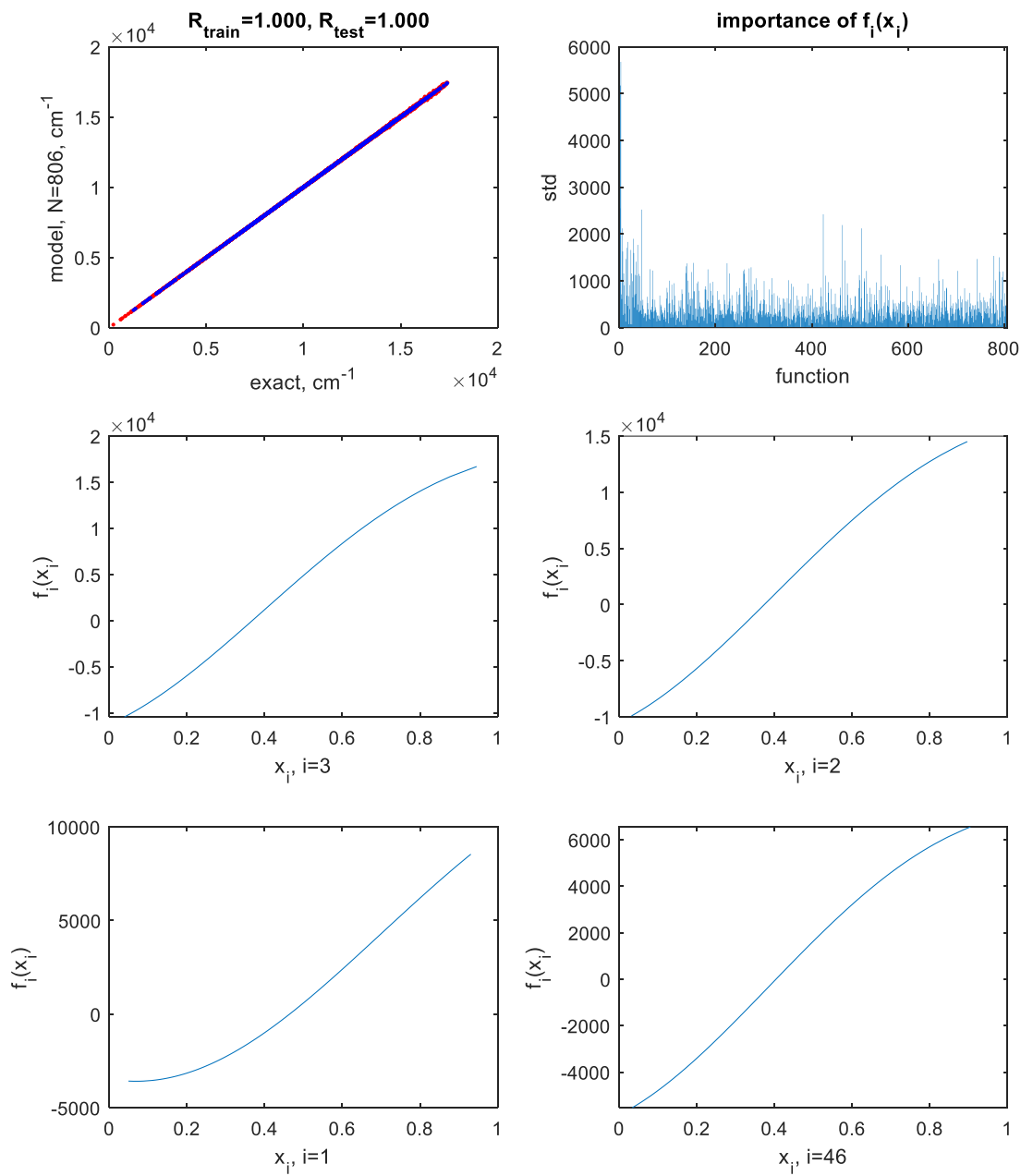

Figure 3. The results of fitting a third-order ($d = 3$) HDMR-NN(GPR) model to the PES of $H_2CO$ with $N$ terms-neurons. Top left: the correlation plot between the exact and fitted values (blue: training set, red: test set). The correlation coefficients for the train and test sets are given on the plot. Top right: the relative importance of terms by standard deviation ("std"). The following three plots are the shapes of the first four most important component functions $f_i(x_i)$.

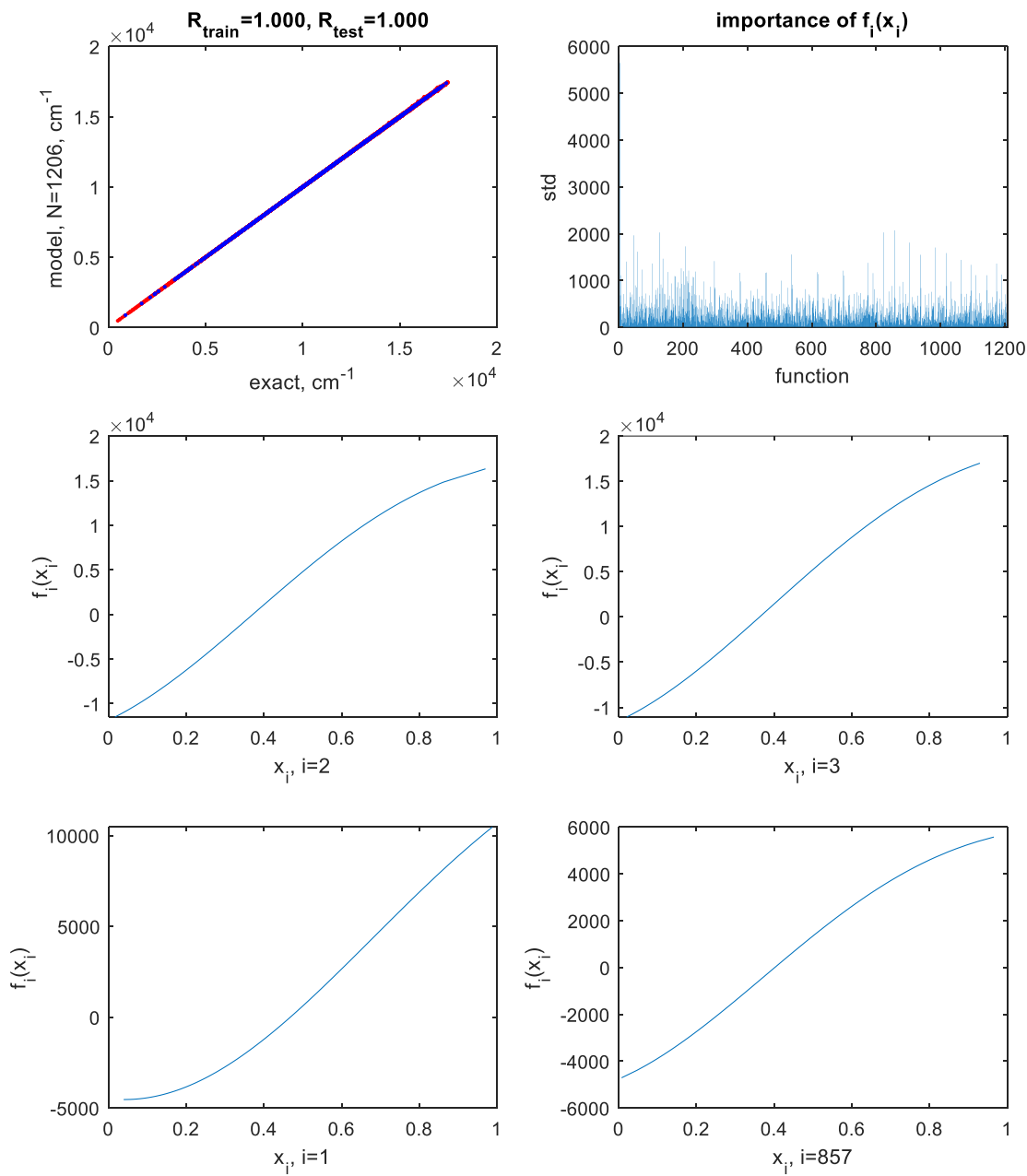

Figure 4. The results of fitting a fourth-order ($d = 4$) HDMR-NN(GPR) model to the PES of $H_2CO$ with $N$ terms-neurons. Top left: the correlation plot between the exact and fitted values (blue: training set, red: test set). The correlation coefficients for the train and test sets are given on the plot. Top right: the relative importance of terms by standard deviation ("std"). The following three plots are the shapes of the first four most important component functions $f_i(x_i)$.

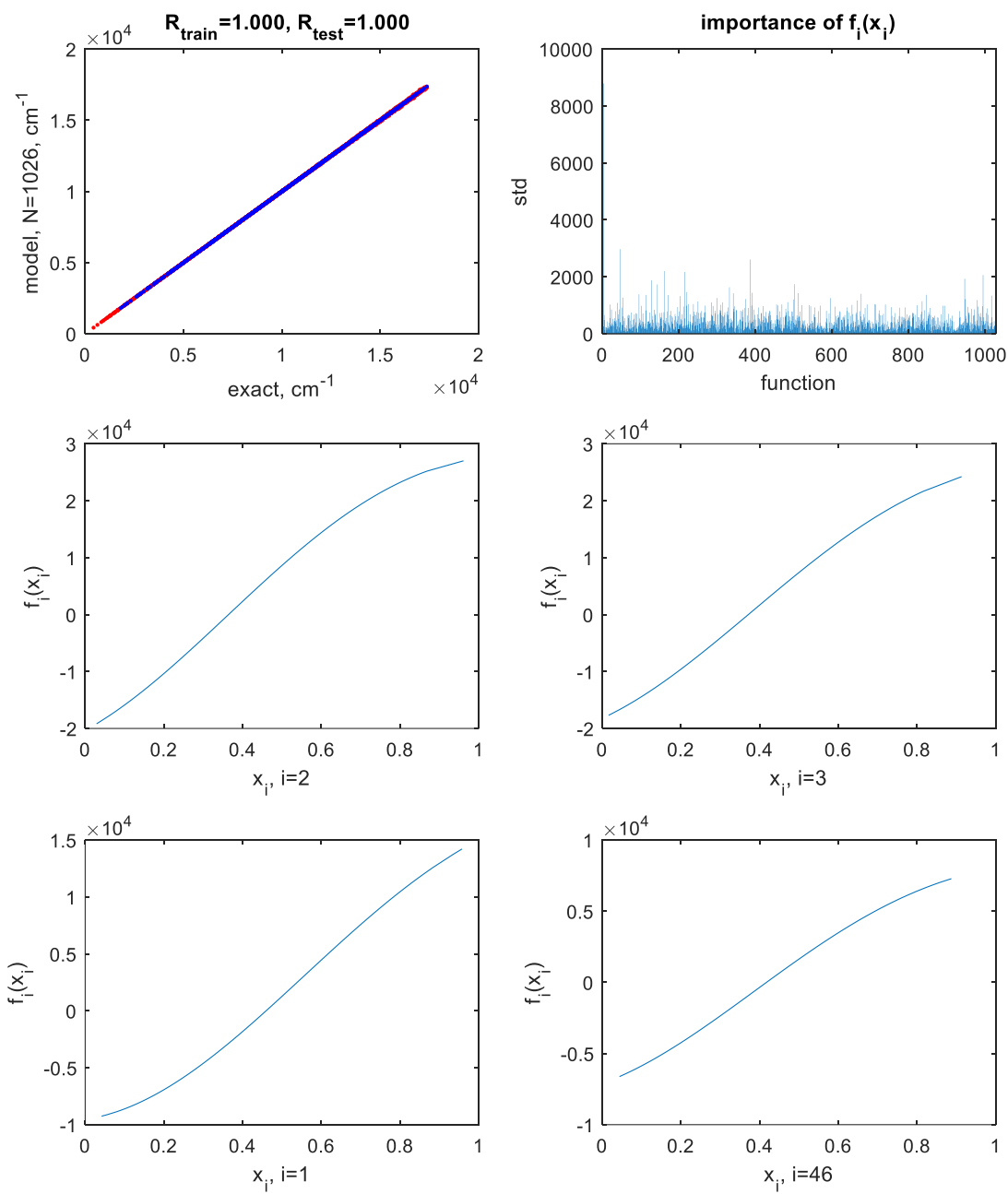

Figure 5. The results of fitting a fifth-order ($d = 5$) HDMR-NN(GPR) model to the PES of $H_2CO$ with $N$ terms-neurons. Top left: the correlation plot between the exact and fitted values (blue: training set, red: test set). The correlation coefficients for the train and test sets are given on the plot. Top right: the relative importance of terms by standard deviation ("std"). The following three plots are the shapes of the first four most important component functions $f_i(x_i)$.

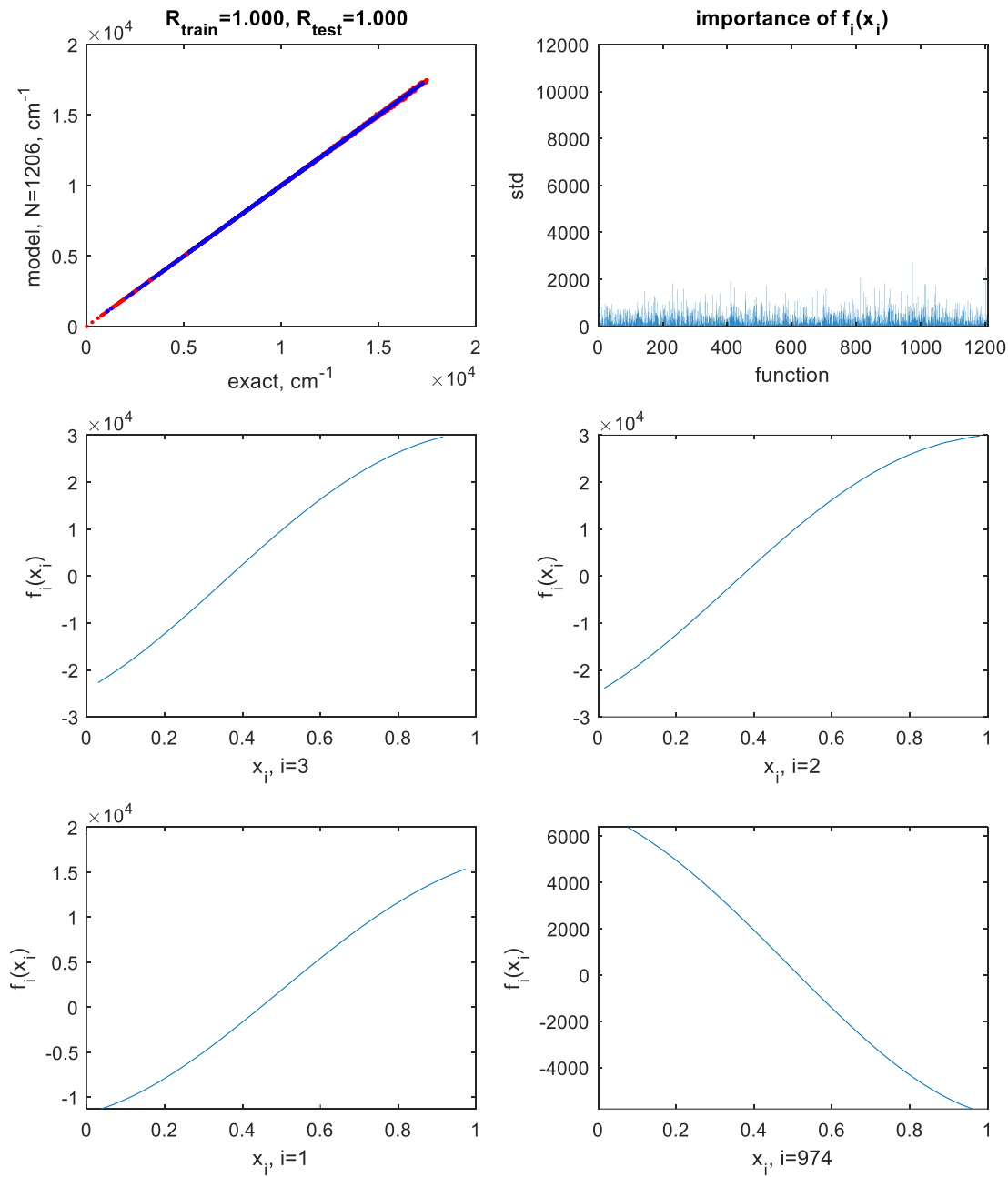

Figure 6. The results of fitting a sixth-order (full dimensional) HDMR-NN(GPR) model to the PES of $H_2CO$ with $N$ terms-neurons. Top left: the correlation plot between the exact and fitted values (blue: training set, red: test set). The correlation coefficients for the train and test sets are given on the plot. Top right: the relative importance of terms by standard deviation ("std"). The following three plots are the shapes of the first four most important component functions $f_i(x_i)$.